\documentclass{article}
\usepackage{longtable}
\usepackage{booktabs}
\usepackage{array}
\usepackage{xcolor}
\usepackage{PRIMEarxiv}
\usepackage{xcolor}
\usepackage{hyperref}
\hypersetup{
    colorlinks=true,
    urlcolor=magenta
}

\usepackage[utf8]{inputenc} 
\usepackage[T1]{fontenc}    
\usepackage{hyperref}       
\usepackage{url}            
\usepackage{booktabs}       
\usepackage{amsfonts}       
\usepackage{nicefrac}       
\usepackage{microtype}      
\usepackage{lipsum}         
\usepackage{fancyhdr}       
\usepackage{graphicx}       
\usepackage{amsmath} 
\usepackage{longtable} 
\usepackage{float} 
\usepackage[skip=2pt]{caption} 
\usepackage{grffile} 
\usepackage{placeins} 

\title{Investigating Thematic Patterns and User Preferences in LLM Interactions using BERTopic
}

\author{
  Abhay Bhandarkar\textsuperscript{*}\\
  Undergraduate\\
  Ramaiah Institute of Technology\\ Bengaluru, India\\
  \texttt{1ms22cs005@msrit.edu}
  \And
  Gaurav Mishra\textsuperscript{*}\\
  Undergraduate\\
  Ramaiah Institute of Technology\\ Bengaluru, India\\
  \texttt{1ms23cs066@msrit.edu}
  \And
  Khushi Juchani\textsuperscript{*}\\
  Lead Data Scientist\\
  Ecofy Finance Private Limited\\ Bengaluru, India\\
  \texttt{khushi.rj1988@gmail.com}
  \And
  Harsh Singhal\textsuperscript{\textdagger}\\
  Visiting Professor\\
  Ramaiah Institute of Technology\\ Bengaluru, India\\
  \texttt{singhalblr@gmail.com}
  \\[1ex]
  \textsuperscript{*}Equal contribution \quad
  \textsuperscript{\textdagger}Guide
}

\begin{document}
\maketitle

\begin{abstract}
This study applies BERTopic, a transformer-based topic modeling technique, to the lmsys-chat-1m dataset, a multilingual conversational corpus built from head-to-head evaluations of large language models (LLMs). Each user prompt is paired with two anonymized LLM responses and a human preference label, used to assess user evaluation of competing model outputs. The main objective is uncovering thematic patterns in these conversations and examining their relation to user preferences, particularly if certain LLMs are consistently preferred within specific topics. A robust preprocessing pipeline was designed for multilingual variation, balancing dialogue turns, and cleaning noisy or redacted data. BERTopic extracted over 29 coherent topics including artificial intelligence, programming, ethics, and cloud infrastructure. We analysed relationships between topics and model preferences to identify trends in model-topic alignment. Visualization techniques included inter-topic distance maps, topic probability distributions, and model-versus-topic matrices. Our findings inform domain-specific fine-tuning and optimization strategies for improving real-world LLM performance and user satisfaction. 
\end{abstract}


\section{Introduction}
\label{sec:introduction}

The proliferation of Large Language Models (LLMs) into diverse applications has underscored the critical importance of understanding human-LLM interaction dynamics. The seminal work "Attention Is All You Need" by Vaswani et al. \cite{vaswani2017attention} introduced the Transformer architecture, a cornerstone for the development of modern LLMs, which have since evolved rapidly, demonstrating remarkable capabilities in natural language understanding and generation. Analysing user interactions with these sophisticated models offers invaluable insights into evolving user behaviours, expectations, and the nuanced levels of trust users place in different LLMs. A comprehensive understanding of the spectrum of user queries—ranging from rudimentary information retrieval to complex problem-solving and creative generation—is essential for the iterative refinement of LLMs. Such understanding not only helps in tailoring LLMs to better cater to user needs but also plays a pivotal role in identifying potential misuse scenarios and enhancing overall AI safety and alignment.

To investigate these interactions, we employed BERTopic for topic modeling on the LMSYS-Chat-1M dataset \cite{zheng2023lmsys}. This dataset was meticulously collected from approximately 210,000 unique IP addresses interacting with the Vicuna demo and the Chatbot Arena website between April and August 2023. Chatbot Arena \cite{chiang2024chatbot} serves as an innovative open platform dedicated to the evaluation of LLMs, primarily leveraging human preference as a key metric. This platform effectively addresses the inherent limitations of static benchmarks by adopting a dynamic, crowdsourced evaluation paradigm. In this setup, users engage in pairwise comparisons of responses generated by different LLMs for the same prompt and cast votes for their preferred response. The platform then employs statistical methodologies, notably the Bradley-Terry model, to efficiently rank the models and estimate their relative performance, complete with confidence intervals. Rigorous data analysis, including preliminary topic modeling efforts, has demonstrated that the platform adeptly captures real-world LLM use cases and successfully differentiates the strengths of various models across a multitude of tasks. Furthermore, Chatbot Arena incorporates sophisticated mechanisms designed to detect and mitigate anomalous user behaviour, thereby ensuring the integrity of the collected preference data. The LMSYS-Chat-1M dataset, born from this initiative, provides a rich tapestry of insights into user interactions with LLMs, rendering it exceptionally valuable for a range of downstream tasks such as content moderation, instruction fine-tuning for improved model alignment, and comprehensive benchmarking. Consequently, we selected this dataset with the objective of identifying conclusive evidence regarding the alignment of top-performing LLMs with specific thematic domains. This was accomplished by systematically extracting models that were majority winners in the preference evaluations and correlating them with the topics of the corresponding user prompts.

Traditionally, topic modeling within the domain of Natural Language Processing (NLP) has been an unsupervised machine learning task aimed at discovering latent thematic structures within a corpus of documents. The core idea is to assign topics to documents based on the co-occurrence patterns of words, effectively summarizing large volumes of text through representative word groups. Prior to the advent of transformer-based techniques, several well-established traditional topic modeling methods gained prominence, including: Latent Semantic Analysis (LSA) \cite{deerwester1990indexing, landauer1997solution}, Probabilistic Latent Semantic Analysis (PLSA) \cite{hofmann1999probabilistic}, Latent Dirichlet Allocation (LDA) \cite{blei2003latent}, and the Correlated Topic Model (CTM) \cite{blei2007correlated}.

However, these traditional topic modeling techniques, often relying on bag-of-words representations, are inherently static and do not adequately accommodate the sequential organization of text within documents. A significant limitation is their inability to effectively capture the rich semantic similarities and contextual nuances present in natural language. While the evolution of NLP did address some of these weaknesses, particularly in methods like LDA, challenges remained. Early neural topic models began to incorporate contextual information to a limited extent and managed vocabulary in continuous embedding spaces, which helped mitigate issues related to rare words or synonyms by clustering them semantically. Nevertheless, these early neural approaches often treated words independently or necessitated custom model training from scratch, limiting their scalability and generalizability. The subsequent and most significant leap in this domain arrived with the development of large pre-trained language models, which provided powerful, context-aware embeddings.

The introduction of LLMs such as BERT (Bidirectional Encoder Representations from Transformers) \cite{devlin2019bert} in 2018 revolutionized the field of NLP by furnishing context-sensitive representations for words and entire documents. BERT demonstrated conclusively that deep bidirectional transformer architectures could capture intricate language nuances, producing embeddings that dynamically account for the surrounding linguistic context. Unlike static word embeddings (e.g., Word2Vec \cite{mikolov2013efficient}, GloVe), which assign a single vector to each word irrespective of its usage, BERT generates distinct embeddings for polysemous words depending on their specific contextual instantiation. Such contextual embeddings have profoundly improved the efficacy of topic modeling by enabling the representation of documents through contextually rich and meaningful vectors, rather than sparse word count matrices. In practice, clustering documents based on these sophisticated embeddings allows for the grouping of semantically similar texts, even if they do not share exact keyword matches, thereby enhancing the interpretability and coherence of the identified topics.

One prominent contextual topic modeling technique that effectivelyleverages BERT embeddings is BERTopic \cite{grootendorst2022bertopic}. Introduced by Grootendorst (2022), BERTopic represents a state-of-the-art approach to unsupervised topic modeling, ingeniously combining transformer-based embeddings with advanced clustering algorithms to generate highly coherent and interpretable topics. BERTopic's modular pipeline leverages pre-trained BERT (or similar transformer) embeddings and a class-based TF-IDF (c-TF-IDF) weighting scheme to create dense document clusters that yield meaningful topic descriptors. Its core process includes:
\begin{itemize}
    \item \textbf{Contextual Embeddings:} Documents are converted into high-dimensional vector representations using transformer models (e.g., MiniLM sentence-transformer embeddings in our case).
    \item \textbf{Dimensionality Reduction:} UMAP (Uniform Manifold Approximation and Projection) \cite{mcinnes2018umap} is applied to project these high-dimensional vectors into lower-dimensional spaces, critically preserving semantic structure while making clustering computationally feasible.
    \item \textbf{Clustering (Topic Formation):} HDBSCAN (Hierarchical Density-Based Spatial Clustering of Applications with Noise) \cite{campello2013density, mcinnes2017hdbscan} is employed to identify groups of similar documents based on their embedding similarities. HDBSCAN has the advantage of automatically determining the optimal number of clusters and effectively labeling outliers as noise.
    \item \textbf{Topic Representation with c-TF-IDF:} A class-based TF-IDF scoring mechanism is computed for keywords within each cluster. This method upweights terms that are frequent in a specific cluster but relatively rare in the overall corpus, thereby deriving descriptive and interpretable labels for each topic.
\end{itemize}

Despite these significant advancements, several challenges persist in the field of topic modeling research and its practical application. Maintaining high topic coherence, determining optimal model parameters (such as the appropriate number of topics for a given corpus), and effective text preprocessing remain considerable hurdles. Suboptimal preprocessing can introduce noise and irrelevant features, whereas overly aggressive filtering might inadvertently remove informative words, thereby complicating the delicate balance required for effective topic modeling.

To address these challenges comprehensively in our study, our methodological approach incorporated several key strategies:
\begin{itemize}
    \item \textbf{Hyperparameter Tuning:} We conducted rigorous experimentation with various hyperparameters, particularly focusing on the number of topics. After extensive evaluation, a configuration yielding 30 topics was selected as optimal for our dataset. During the clustering phase, documents that HDBSCAN labeled as '-1' (i.e., uncategorized outliers) were systematically excluded from the primary topic analysis to enhance the overall coherence and interpretability of the derived topics.
    \item \textbf{Extensive Text Preprocessing:} A meticulous text preprocessing pipeline was implemented. This involved the removal of non-English text segments, emojis, URLs, non-alphanumeric characters, and a set of custom-defined "stop prompts" (standardized instructional phrases). These steps were crucial for maintaining a corpus of uniform and semantically meaningful tokens.
    \item \textbf{Experiments with Different Pipelines:} Prior to settling on our final configuration, we tested various alternative pipelines. This included experimenting with different sentence embeddings (e.g., other variants beyond MiniLM), alternative clustering algorithms such as KMeans, and using CountVectorizer for feature extraction. These preliminary experiments provided valuable baseline insights and served to justify our final methodological choices.
    \item \textbf{Final Adoption of BERTopic Configuration:} After extensive comparative evaluation, the BERTopic framework, combined with CountVectorizer for generating initial document-term matrices (often used internally by BERTopic or as a precursor if not using direct embeddings for certain steps), was determined to provide the most coherent and interpretable topics for the LMSYS-Chat-1M dataset.
\end{itemize}
This research explores the application and systematic optimization of BERTopic for analyzing large-scale human-LLM conversational data. It assesses its efficacy in comparison to traditional topic modeling methods and highlights specific refinements aimed at enhancing the interpretability of results and improving computational efficiency in handling such complex datasets.

\section{Literature Review}
\label{sec:literature_review}

The evolution of topic modeling techniques has witnessed remarkable progress over the past decades, transitioning from traditional probabilistic models to sophisticated deep neural architectures. Early foundational work centered on statistical methods such as Latent Dirichlet Allocation (LDA), which conceptualizes a document as a mixture of topics and, reciprocally, topics as a mixture of words \cite{blei2003latent}. Concurrently, Probabilistic Latent Semantic Indexing (PLSI), also known as Probabilistic Latent Semantic Analysis (PLSA), proposed a latent class model for detecting textual themes but exhibited several shortcomings, notably a propensity for overfitting and the lack of a well-defined generative model for producing new documents \cite{hofmann1999probabilistic}. These traditional methods were predominantly built upon bag-of-words (BoW) representations, a paradigm that inherently fails to capture word order and contextual information, thereby constraining their performance on modern, context-sensitive language datasets.

The advent of transformer-based models heralded a paradigm shift, fundamentally revolutionizing natural language processing by enabling the generation of context-sensitive embeddings that adeptly capture subtle semantic distinctions, such as polysemy and complex syntactic relationships. BERT (Bidirectional Encoder Representations from Transformers) played a pivotal role in this transformation by introducing bidirectional conditioning across all layers of the neural network, thereby redefining the quality and richness of semantic representations \cite{devlin2019bert}. Such advancements paved the way for the development of neural topic models that operate directly within dense vector spaces, a significant departure from traditional methods reliant on sparse word frequency matrices.

BERTopic \cite{grootendorst2022bertopic} stands as a quintessential example of these modern methods, employing a modular pipeline that typically consists of three distinct stages: first, UMAP (Uniform Manifold Approximation and Projection) \cite{mcinnes2018umap, mittal2024dimensionality} facilitates the embedding of high-dimensional sentence embeddings into a lower-dimensional space—a process of dimensionality reduction—while preserving both global and local semantic relationships with higher fidelity than competing techniques such as t-SNE \cite{van2008visualizing}. Comparative assessments across numerous domains have demonstrated the utility of UMAP's dimension reduction characteristics in managing complex linguistic data \cite{mittal2024dimensionality}. Second, HDBSCAN (Hierarchical Density-Based Spatial Clustering of Applications with Noise) \cite{campello2013density, mcinnes2017hdbscan, asyaky2021improving} identifies dense clusters within the reduced embedding space without necessitating a pre-specification of the number of topics, exhibiting optimal efficiency when applied to noisy, real-world datasets. Third, a class-based TF-IDF (c-TF-IDF) mechanism is utilized to identify discriminating terms for each cluster, leading to highly interpretable topic representations \cite{grootendorst2022bertopic}.

The improvement in topic clustering quality has been significantly propelled by the use of sentence-level representations, such as those generated by Sentence-BERT (SBERT) \cite{reimers2019sentence}. SBERT captures dense sentence meaning, enabling the formation of more coherent topic clusters than conventional word-level vector methods, like those based on Word2Vec \cite{mikolov2013efficient}. When compared directly to other neural clustering methods, such as Top2Vec \cite{angelov2020top2vec}, BERTopic has consistently demonstrated superior performance in creating well-defined topic boundaries and adapting to diverse domains. Recent advances have sought to extend these capabilities further, with researchers exploring hybrid models that combine contextualized topic models with advanced embeddings like MPNet to analyze user feedback and other complex textual data \cite{asnawi2023combination}.

The framework for evaluating topic quality has evolved beyond traditional intrinsic measures like perplexity to incorporate coherence scores and human-rated assessments. Lau et al. \cite{lau2014machine} proposed automatic evaluation techniques that quantify semantic coherence by capturing word co-occurrence patterns, demonstrating a high correlation with human judgments of topic quality. Korenčić et al. \cite{korencic2021topic} further enhanced evaluation toolkits by introducing topic coverage measures that assess the breadth and depth of the identified themes. Such robust evaluation tools are particularly crucial when analyzing the output of LLMs, where creative linguistic variations and subtle thematic shifts can misleadingly affect probability-based evaluation metrics.

Topic modeling has found extensive applications across a wide array of fields. For instance, Gürcan \cite{gurcan2018major} utilized probabilistic topic modeling to analyze trends in big data research between 2013 and 2017, revealing broad shifts in thematic focus. Similarly, Johri et al. \cite{johri2011utilizing} applied topic modeling techniques to identify emerging research trends within engineering education. Du et al. \cite{du2014topic} investigated topic models that explicitly incorporate ordering regularities for the purpose of topic segmentation, while Yang et al. \cite{yang2013line} introduced an adaptive topic evolution model tailored for web discussion contexts. Such specific applications underscore the versatility and adaptability of topic modeling techniques in diverse research and practical settings. Other relevant work includes the comparison of statistical models for topic discovery in specific languages like Urdu \cite{mustafa2023discovering}.

More recent initiatives, such as Chatbot Arena \cite{chiang2024chatbot}, provide valuable human preference baselines for LLMs but often lack systematic, topic-level analyses of the thematic composition inherent in model outputs. The LMSYS-Chat-1M corpus \cite{zheng2023lmsys} contributes significantly to this foundation by offering access to one million real-world LLM conversations, yet comprehensive topic-level analysis of this resource has been limited. Applying advanced topic modeling methods like BERTopic to these rich corpora holds the potential to uncover how different LLMs organize knowledge into coherent topics. Furthermore, such analyses may reveal inherent biases, patterns in response diversity, and temporal shifts in thematic focus, thereby enriching existing LLM evaluation frameworks and guiding future development.

\section{Methodology}
\label{sec:methodology}

\begin{figure}[t!] 
    \centering
    \includegraphics[width=1.05\linewidth]{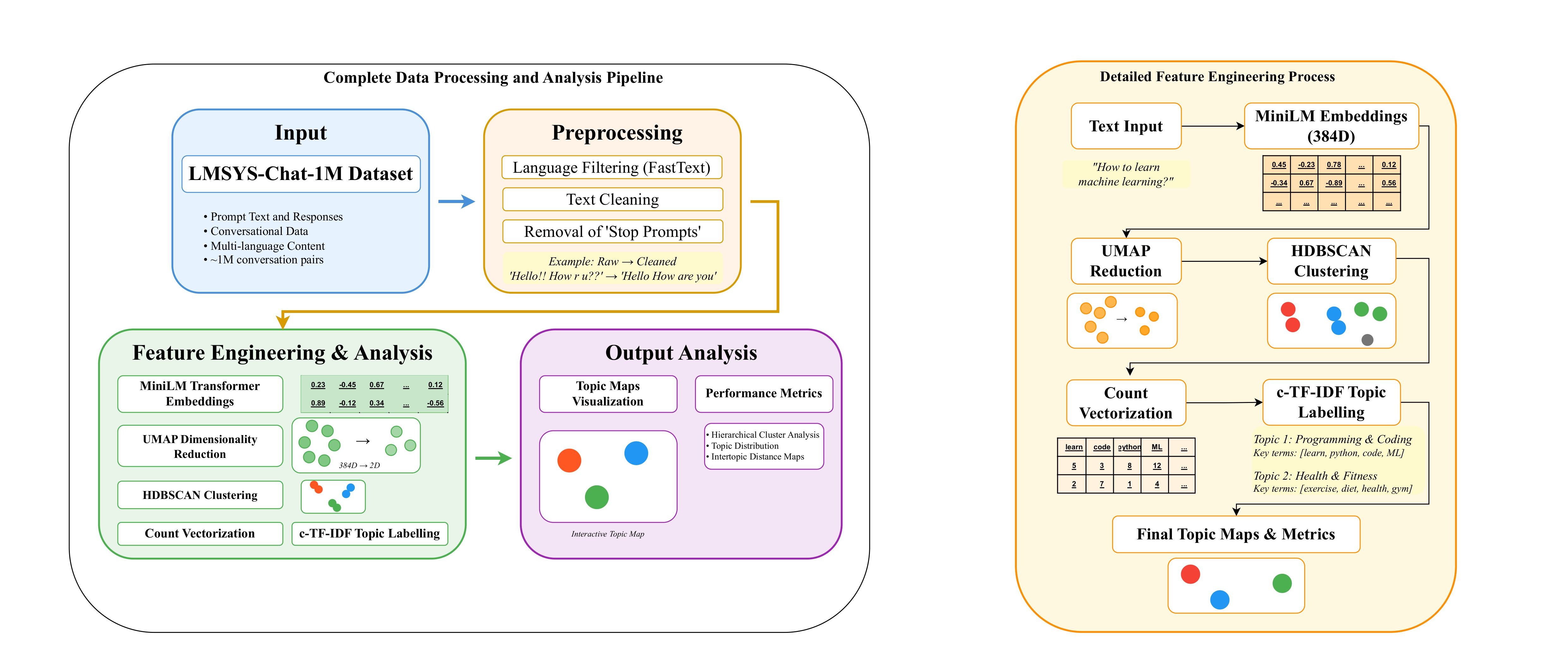} 
    \caption{Architectural diagram of the topic modeling pipeline.}
    \label{fig:architectural_diagram}
\end{figure}

Our approach employs BERTopic for unsupervised topic modeling on the LMSYS-Chat-1M dataset, specifically collected through human-Large Language Model (LLM) interactions. This section details the systematic methodology, encompassing dataset acquisition and characterization, preliminary data exploration, a rigorous multi-stage data preprocessing pipeline, contextual topic modeling with BERTopic, meticulous hyperparameter optimization, and comparative validation. The ultimate aim is to discern semantically coherent topics and analyze associated human preferences. The complete pipeline is depicted in the architectural diagram (Figure \ref{fig:architectural_diagram}).

\subsection{Dataset Acquisition and Characterization}
The empirical foundation of this research is the LMSYS-Chat-1M dataset \cite{zheng2023lmsys}, collected between April and August 2023 via the Vicuna demo and Chatbot Arena, involving approximately 210,000 unique IP addresses. This multilingual corpus includes pairwise comparisons between various LLM responses to user-generated prompts, with human-preference labels used to rank model responses efficiently. From the dataset, each conversation (consisting of user prompts and LLM responses) was treated as a single \textit{document} for modeling purposes. These human preference labels ($P_L$) constitute a critical component for assessing the relative performance of different LLMs across various conversational contexts.

\subsection{Preliminary Data Exploration}
\begin{figure}[htbp]
  \centering
  \includegraphics[width=\textwidth]{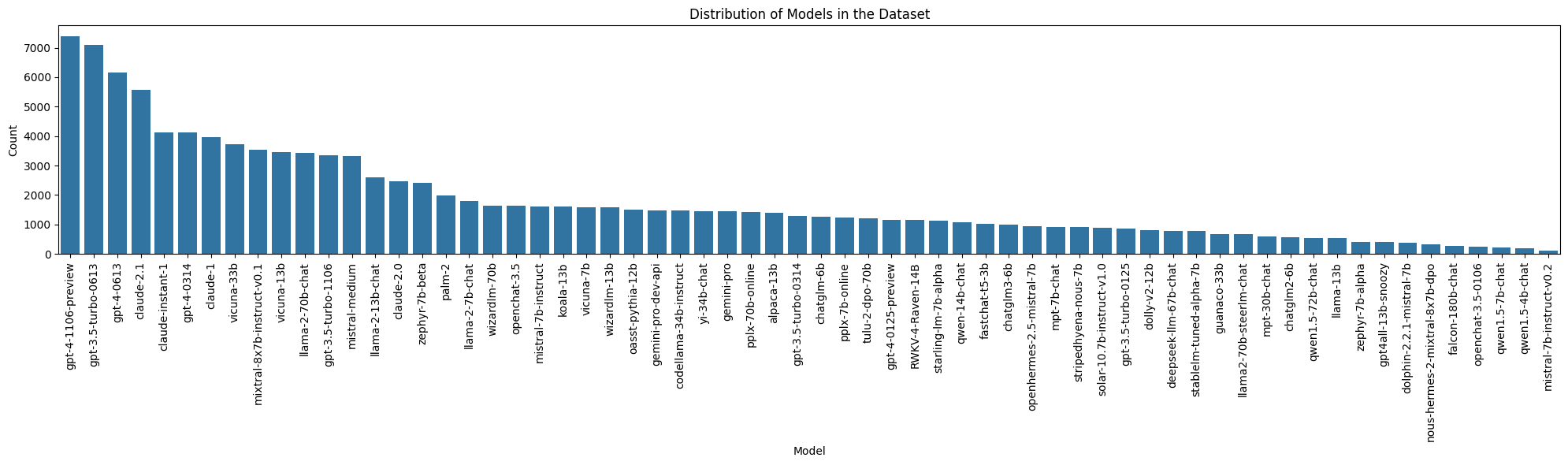}
  \caption{Distribution of LLM appearances in the LMSYS-Chat-1M dataset}
  \label{fig:eda_model_bar_distribution}
\end{figure}

\begin{figure}[t] 
    \centering
    \begin{minipage}{0.48\textwidth}
        \centering
        \includegraphics[width=\linewidth]{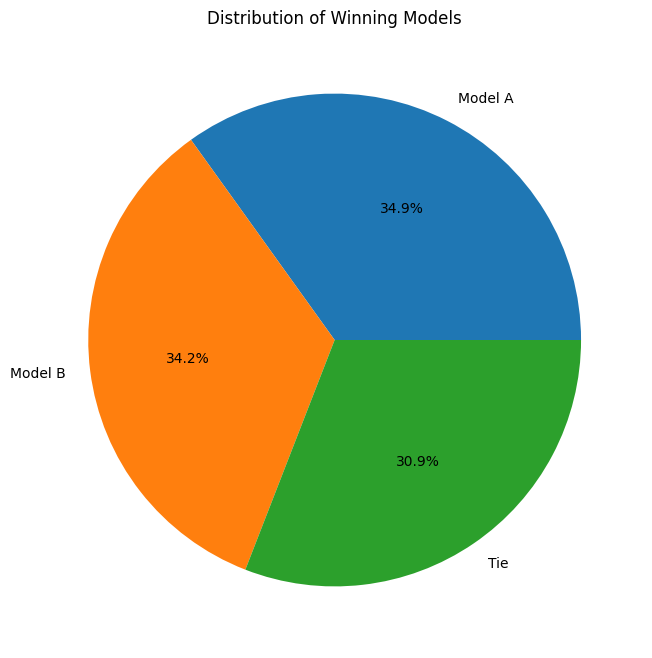} 
        \caption{Overall win/loss/tie distribution in pairwise evaluations (34.9\%/34.2\%/30.9\% split).}
        \label{fig:eda_win_tie_distribution}
    \end{minipage}\hfill 
    \begin{minipage}{0.48\textwidth}
        \centering
        \includegraphics[width=\linewidth]{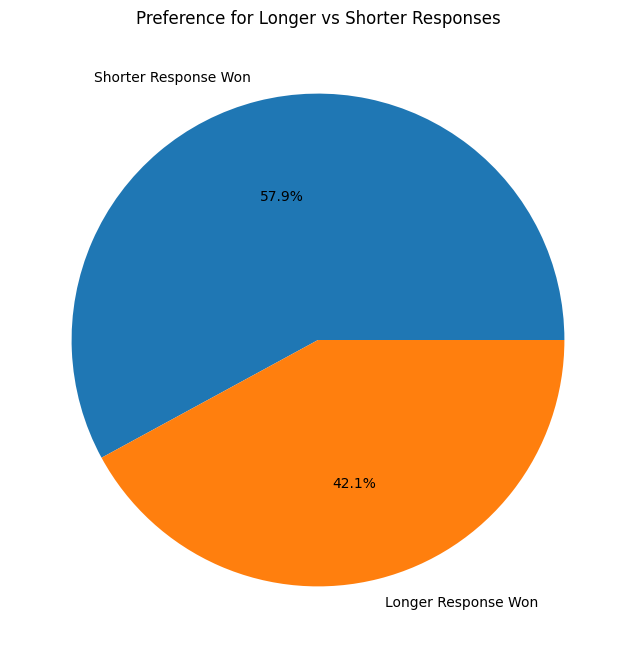} 
        \caption{User preference for shorter vs. longer responses (57.9\%/42.1\% split).}
        \label{fig:eda_response_length_preference}
    \end{minipage}
    \label{fig:side_by_side_eda} 
\end{figure}

Before delving into topic modeling, a high-level exploration of the raw conversation data was performed to understand model representation and basic user preferences within the LMSYS-Chat-1M corpus.
\begin{itemize}
    \item \textbf{Model Appearance Frequency:} We tallied the number of times each LLM appeared as a respondent. As shown in Figure \ref{fig:eda_model_bar_distribution}, a small set of models (e.g., \texttt{gpt-4-1106-preview}, \texttt{gpt-3.5-turbo-0613}, \texttt{claude-2.1}) dominate the dataset, each contributing over 5,000 prompts, while dozens of other models appear far less frequently.
    \item \textbf{Win/Loss/Tie Distribution:} For all pairwise evaluations where Model A is compared against Model B, Figure \ref{fig:eda_win_tie_distribution} illustrates that wins and ties are roughly equally distributed—Model A wins 34.9\% of comparisons, Model B wins 34.2\%, and ties occur 30.9\% of the time. This near-uniform split indicates no single model overwhelmingly outperforms its competitor at a corpus-wide level.
    \item \textbf{Response Length Preference:} To investigate if response length influences human preference, we marked each winning response as either the shorter or longer of the two model outputs. As Figure \ref{fig:eda_response_length_preference} shows, shorter responses win 57.9\% of the time compared to 42.1\% for longer ones, suggesting a general user tendency to favor more concise answers from LLMs.
\end{itemize}

\subsection{Data Preprocessing}
To ensure the fidelity and relevance of the input data for subsequent topic modeling, a comprehensive preprocessing protocol was instituted. This protocol was designed to normalize textual data, mitigate noise, and enhance the semantic signal.
\begin{itemize}
    \item \textbf{Language Filtration:} The analysis was confined to English-language interactions. FastText \cite{joulin2017bag}, a library for efficient text classification and representation learning, was utilized to programmatically identify and isolate English-language segments. All non-English content was systematically excluded.
    \item \textbf{Text Normalization and Cleaning:} Standard text sanitization procedures were executed using regular expressions to remove unwanted textual noise. Non-ASCII characters (e.g., emojis, emoticons) were stripped out. The text was processed to ensure an ASCII-only string with normalized spacing, and without special characters, backslashes, or excessive formatting whitespace. This also involved removing Uniform Resource Locators (URLs).
    \item \textbf{"Stop Prompt" Removal Consideration:} We experimented with cleaning records containing gibberish prompts or specific recurring boilerplate text from chat prompts (termed "stop-prompts") that were deemed potentially irrelevant for topic identification. However, this step was later discarded as it did not yield a significant impact on the final topics' overall distribution against winning models.
\end{itemize}
Once a cleaned and preprocessed dataset was generated, it was used as input for the BERTopic models.

\subsection{Topic Modeling with BERTopic}
BERTopic \cite{grootendorst2022bertopic}, a state-of-the-art unsupervised topic modeling technique, was selected for its capacity to leverage contextual embeddings and generate semantically meaningful topics. The BERTopic pipeline, as implemented in this study, comprises the following sequential stages:

\subsubsection{Document Embedding Generation}
Each preprocessed textual document $d_i$ (representing entire conversations) was transformed into a high-dimensional dense vector embedding $e_i$. This transformation was primarily achieved using the \texttt{all-MiniLM-L6-v2} sentence-transformer model (related to work by Reimers and Gurevych \cite{reimers2019sentence}), chosen for its effectiveness in capturing semantic similarity. Each embedding $e_i$ resides in a vector space $\mathbb{R}^D$, where $D=384$ for \texttt{all-MiniLM-L6-v2}. Experiments were also performed with other embeddings like \texttt{all-mpnet-base-v2}; however, these were not as effective for topic segregation, often resulting in clustered topics with negligible coherence.

\subsubsection{Dimensionality Reduction with UMAP}
To address the "curse of dimensionality" inherent in raw transformer embeddings, Uniform Manifold Approximation and Projection (UMAP) \cite{mcinnes2018umap} was applied as the default dimensionality reduction technique. UMAP constructs a weighted graph by assigning edge weights $w_{ij}$ between points $x_i$ and $x_j$ using the function:
\[ w_{ij} = \exp\left( -\frac{\max(0, d(x_i, x_j) - \rho_i)}{\sigma_i} \right) \]
where $d(x_i, x_j)$ is the distance between the points, $\rho_i$ represents the distance to the nearest neighbor of $x_i$, and $\sigma_i$ is a local normalization factor. This formulation enables UMAP to capture the manifold structure of the data, which is then optimized in a lower-dimensional space $e'_i \in \mathbb{R}^d$ (where $d \ll D$) to preserve local connectivity. This compact representation both speeds up clustering and enhances the quality of the discovered topics, serving as an intermediate process between the transformer encoder and the clustering algorithm.

\subsubsection{Clustering with HDBSCAN}
The lower-dimensional embeddings $e'_i$ were clustered using Hierarchical Density-Based Spatial Clustering of Applications with Noise (HDBSCAN) \cite{campello2013density, mcinnes2017hdbscan}. HDBSCAN is a robust algorithm that discovers clusters of arbitrary shape and varying densities without requiring a pre-specified number of clusters. It defines clusters as dense regions separated by areas of lower density.

HDBSCAN begins by computing a distance matrix (using Euclidean distance on UMAP embeddings) and transforms these into a density-aware mutual reachability distance. The mutual reachability distance $d_{\text{mreach}}(a, b)$ between two points $a$ and $b$ is defined as:
\[ d_{\text{mreach}}(a, b) = \max\left\{ d(a, b), \text{core}_k(a), \text{core}_k(b) \right\} \]
where $d(a,b)$ is the base Euclidean distance, and $\text{core}_k(x)$ is the distance from point $x$ to its $k$-th nearest neighbor, determined by the \texttt{min\_samples} parameter (often referred to as `minPts` in literature). This prevents sparse points from incorrectly linking dense clusters. 

Using these distances, HDBSCAN builds a minimum spanning tree and then a hierarchical cluster tree (dendrogram). It extracts a flat clustering by selecting clusters based on their stability—clusters persisting over a wide range of density thresholds are favored. We tuned HDBSCAN’s parameters, setting \texttt{min\_cluster\_size} substantially higher than default values to avoid over-fragmentation, ensuring topics were broad and meaningful. The \texttt{min\_samples} parameter was also adjusted to balance sensitivity in topic detection with noise filtering. Points not belonging to any dense region were labeled as noise (cluster ID -1). This capacity to label outliers is valuable as it avoids forced assignments, preserving topic coherence. A comparative analysis using KMeans was also performed; however, KMeans fell short due to its inability to handle noise effectively and its requirement to pre-specify $k$, often forcing data points into ill-fitting clusters. Thus, HDBSCAN was finalized as our clustering method.
Figure 4 shows method of clustering by employing HDBSCAN.
\begin{figure}[htbp]
    \centering
    \includegraphics[width=1\linewidth]{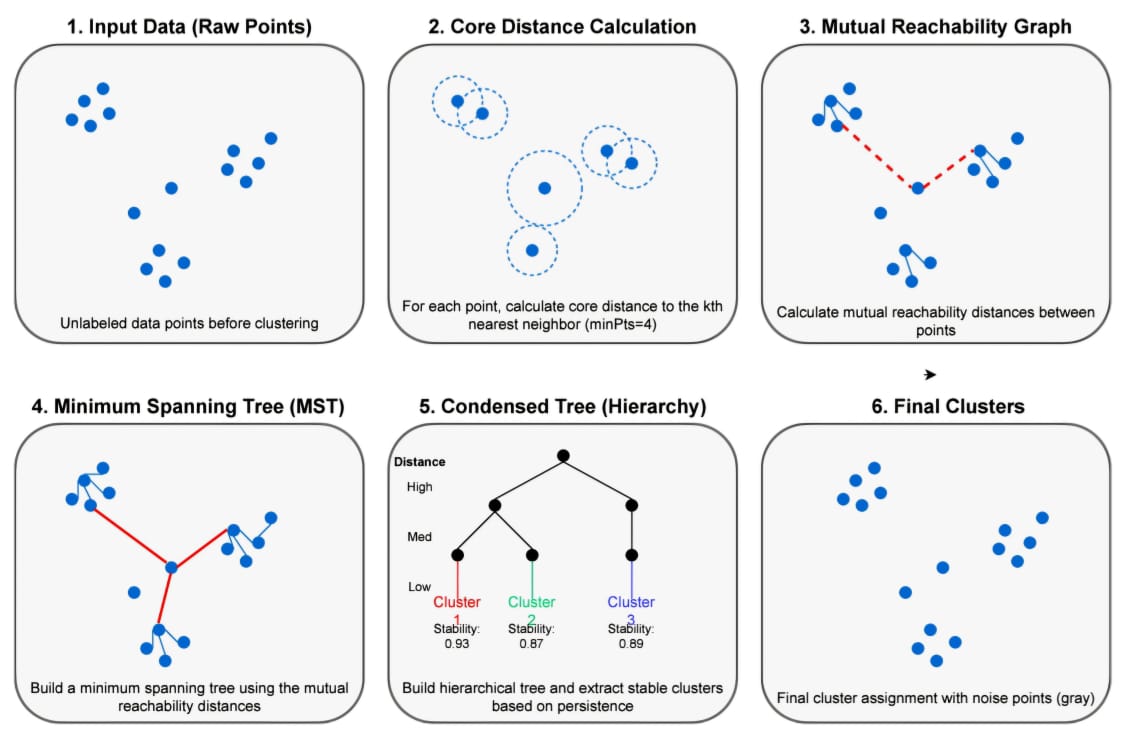}
    \caption{Clustering through HDBSCAN}
    \label{fig:hdbscan_methodology}
\end{figure}

\subsubsection{Topic Representation using CountVectorizer and c-TF-IDF}
Once document clusters were identified, topic representations were generated. We used \texttt{CountVectorizer} to tokenize the documents within each cluster, applying standard English stop words to filter out non-informative terms. This step does not affect the quality of the clusters, which are formed prior to this representation stage.

Subsequently, BERTopic’s class-based TF–IDF (c-TF-IDF) approach was utilized to identify the most important terms for each topic \cite{grootendorst2022bertopic}. The c-TF-IDF method treats all documents within a given cluster as a single consolidated document representing that topic. A TF–IDF-like score is then computed for words in each of these "topic documents" relative to the collection of all such topic documents.

Formally, let $C$ be the number of clusters (topics) excluding noise, and $f_{w,c}$ be the frequency of word $w$ in cluster $c$. The class-term frequency $\text{TF}_{c,w}$ is:
\[ \text{TF}_{c,w} = \frac{f_{w,c}}{\sum_{u \in V} f_{u,c}} \]
where $V$ is the vocabulary and the denominator is the total number of words in cluster $c$.
Let $f_w = \sum_{c'=1}^{C} f_{w,c'}$ be the total frequency of word $w$ across all clusters, and $A = \frac{1}{C} \sum_{c'=1}^{C} \sum_{u \in V} f_{u,c'}$ be the average number of words per cluster. The class-based inverse document frequency for word $w$ is:
\[ \text{IDF}_w^{(\text{c-TF-IDF})} = \log\left(1 + \frac{A}{f_w}\right) \]
The c-TF-IDF score $s_{c,w}$ for word $w$ in topic $c$ is:
\[ s_{c,w} = \text{TF}_{c,w} \times \text{IDF}_w^{(\text{c-TF-IDF})} \]
Terms with the highest $s_{c,w}$ scores for each topic were selected as its representative keywords, enabling the assignment of interpretable, human-readable labels.

\subsection{Noise Handling}
In our analysis, points labeled as noise (cluster ID -1) by HDBSCAN were treated as uncategorized conversations and excluded from the main topic interpretation and visualization. This ensures that the defined topics remain coherent. While the proportion of noise points was noted, they were not the focus of the thematic analysis.

\subsection{Hyperparameter Optimization}
Optimal performance of the BERTopic model was pursued through rigorous experimentation with its key hyperparameters. A primary focus was the \texttt{min\_topic\_size} for HDBSCAN (which influences \texttt{min\_cluster\_size}), and UMAP parameters like \texttt{n\_neighbors}. Iterative adjustments and qualitative evaluations of topic coherence led to the selection of a configuration that yielded 29 distinct, semantically coherent topics (after initially targeting 30 and accounting for the outlier topic). Documents classified as outliers (Topic -1) were deliberately excluded from the primary thematic analysis to preserve the integrity and interpretability of the core topics.

\subsection{Comparative Validation}
To ascertain the robustness and superiority of the chosen BERTopic-based methodology, comparative analyses were conducted:
\begin{itemize}
    \item \textbf{Traditional Latent Semantic Models:} Baselines included Latent Semantic Analysis (LSA) \cite{deerwester1990indexing}, Latent Dirichlet Allocation (LDA) \cite{blei2003latent}, and Probabilistic Latent Semantic Analysis (PLSA) \cite{hofmann1999probabilistic}. Qualitative assessments of topic coherence and semantic interpretability indicated that BERTopic surpassed these methods for this dataset.
    \item \textbf{Alternative Embedding and Clustering Strategies:} As mentioned, alternative embeddings (e.g., \texttt{all-mpnet-base-v2}) and clustering algorithms (KMeans) were evaluated. These experiments consistently reaffirmed the efficacy of the selected \texttt{all-MiniLM-L6-v2} embeddings and HDBSCAN clustering configuration for this specific analytical task, primarily due to superior topic coherence and effective noise handling.
\end{itemize}

\subsection{Visualization of Topic Clusters (Conceptual)}
A standard output for interrogating the BERTopic model is the visualization of topic clusters in a lower-dimensional space. Typically, this involves plotting the 2D or 3D UMAP-reduced document embeddings, where each point is colored according to its HDBSCAN-assigned topic ID. Such a visualization (not presented pictorially in this manuscript but conceptually part of the BERTopic workflow) allows for a qualitative assessment of topic separability, density, and inter-topic relationships, as well as the identification of outlier distributions.

\section{Results and Discussion}
\label{sec:results_discussion}

This section presents the principal findings derived from the application of the BERTopic modeling pipeline to the LMSYS-Chat-1M dataset. It encompasses the characterization of the identified topics, their prevalence and distribution, an analysis of cumulative topic coverage, and a detailed examination of Large Language Model (LLM) performance across these thematic categories, based on human preference data.

\subsection{Identified Thematic Clusters}
HDBSCAN identified clusters of similar documents, automatically determining optimal clusters and classifying outliers as noise. We complemented HDBSCAN’s density‐based topics with a simple agglomerative dendrogram over the topic centroids (e.g. their c-TF-IDF vectors). In this bottom-up tree, each topic starts alone and the two closest clusters (by average-linkage distance) merge stepwise—so topics that join low on the horizontal axis are very similar, while those merging further right are more distinct. Cutting the tree at a chosen distance reveals natural “super-clusters” (e.g. politics and welfare topics vs. cooking or technical ones), giving an intuitive semantic map of how our BERTopic topics relate .

\begin{figure}[htbp]
    \centering
    \includegraphics[width=1\linewidth]{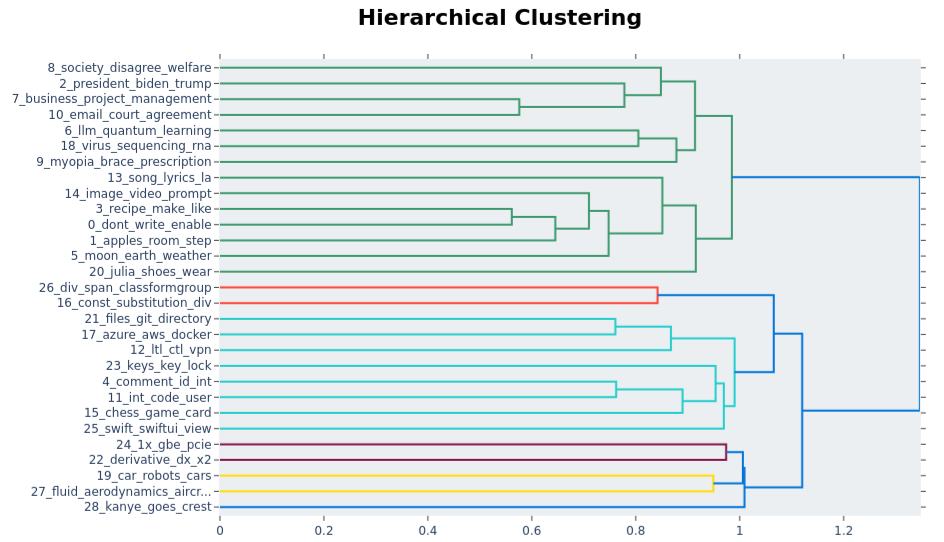}
    \caption{Dendrogram of topic hierarchy.}
    \label{fig:topic_hierarchy}
\end{figure}

HDBSCAN constructs a condensed cluster tree from the minimum spanning tree of mutual reachability distances, tracking the birth and death of clusters as the density threshold ($\lambda$) varies. The stability of each cluster—quantified by its persistence across ($\lambda$)—determines the final flat clustering. By coupling this density-driven view with an agglomerative Dendrogram (e.g., via hierarchical topic modeling), we validate that the resulting topics emerge from robust, high-density regions while also uncovering their higher-order semantic relationships. This dual perspective enhances interpretability and can guide downstream topic merging or labelling.

The BERTopic model successfully identified 29 distinct thematic clusters (Topics 0 through 28), excluding the outlier category (Topic -1), from the dataset. These topics span a diverse spectrum of subjects prevalent in human-LLM dialogues. The top keywords for each topic, derived using c-TF-IDF, were reviewed to assign human-understandable labels. Table \ref{tab:identified_topics} enumerates these topics along with their descriptive labels, inferred from their most representative c-TF-IDF keywords and validated through inspection of constituent exemplar prompts.

\begin{longtable}{@{}ll@{}}
\caption{Identified Topics and Corresponding Descriptions from BERTopic Model} \label{tab:identified_topics} \\
\toprule
\textbf{Topic ID} & \textbf{Description} \\
\midrule
\endfirsthead
\multicolumn{2}{c}%
{{\bfseries \tablename\ \thetable{} -- continued from previous page}} \\
\toprule
\textbf{Topic ID} & \textbf{Description} \\
\midrule
\endhead
\midrule
\multicolumn{2}{r}{{Continued on next page}} \\
\midrule
\endfoot
\bottomrule
\endlastfoot
Topic 0  & Gaming and user-assistant interaction      \\
Topic 1  & Cognitive Trick Problems / Logic Puzzles     \\
Topic 2  & Politics, Celebrities, and Current Events    \\
Topic 3  & Cooking, Recipes, and Event Planning       \\
Topic 4  & Programming, SQL, RDBMS, Database          \\
Topic 5  & Science, Astronomy, Astrophysics Queries     \\
Topic 6  & Machine Learning and Advanced AI Concepts    \\
Topic 7  & Finance, Business, Economic Strategies     \\
Topic 8  & Social Issues and Ethical Dilemmas         \\
Topic 9  & Health Advice and Medical Concerns         \\
Topic 10 & Creative Content Creation, Writing, Email Communication \\
Topic 11 & Programming Concepts and Software Development \\
Topic 12 & Technology Recommendations and Comparisons   \\
Topic 13 & Song, Lyrics, Creative Writing             \\
Topic 14 & Media Editing and Technical Support Queries  \\
Topic 15 & Math Problems and Logic Games              \\
Topic 16 & JavaScript, React, Web Development         \\
Topic 17 & Cloud Infrastructure and Kubernetes Management\\
Topic 18 & Genetics, COVID-19, Biological Sciences     \\
Topic 19 & Automobiles, Engineering, Transportation Queries\\
Topic 20 & Fashion Advice and Clothing Queries        \\
Topic 21 & Git, Linux, OS, Automation, DevOps Solutions \\
Topic 22 & Advanced Calculus and Mathematical Theorems    \\
Topic 23 & Keyboard Inputs and Text Entry Optimization    \\
Topic 24 & Linux Storage Management and NAS Solutions   \\
Topic 25 & Swift Programming and SwiftUI Development    \\
Topic 26 & HTML Forms and Web Interface Customization   \\
Topic 27 & Aerodynamics and Fluid Dynamics Principles   \\
Topic 28 & Singers, Creative Writing, Rhyming Narratives \\
\end{longtable}

\subsection{Topic Distribution and Coverage}
The analysis of topic distribution revealed heterogeneity in the volume of prompts associated with each identified topic. Certain topics aggregated thousands of user prompts, indicating areas of broad and significant user interest, whereas other topics represented more specialized or niche subjects, characterized by a comparatively smaller number of associated interactions. A notable observation was that a substantial number of prompts (approximately 24,000) were categorized by HDBSCAN into the outlier topic (Topic -1). This suggests a high degree of lexical and semantic diversity in these particular queries, which precluded their coherent assignment to one of the 29 core thematic clusters. Such information is useful to understand how broad or niche each topic is within the dataset.

\subsubsection{Cumulative Topic Coverage Analysis}
\begin{figure}[htbp] 
    \centering
    \includegraphics[width=0.8\textwidth]{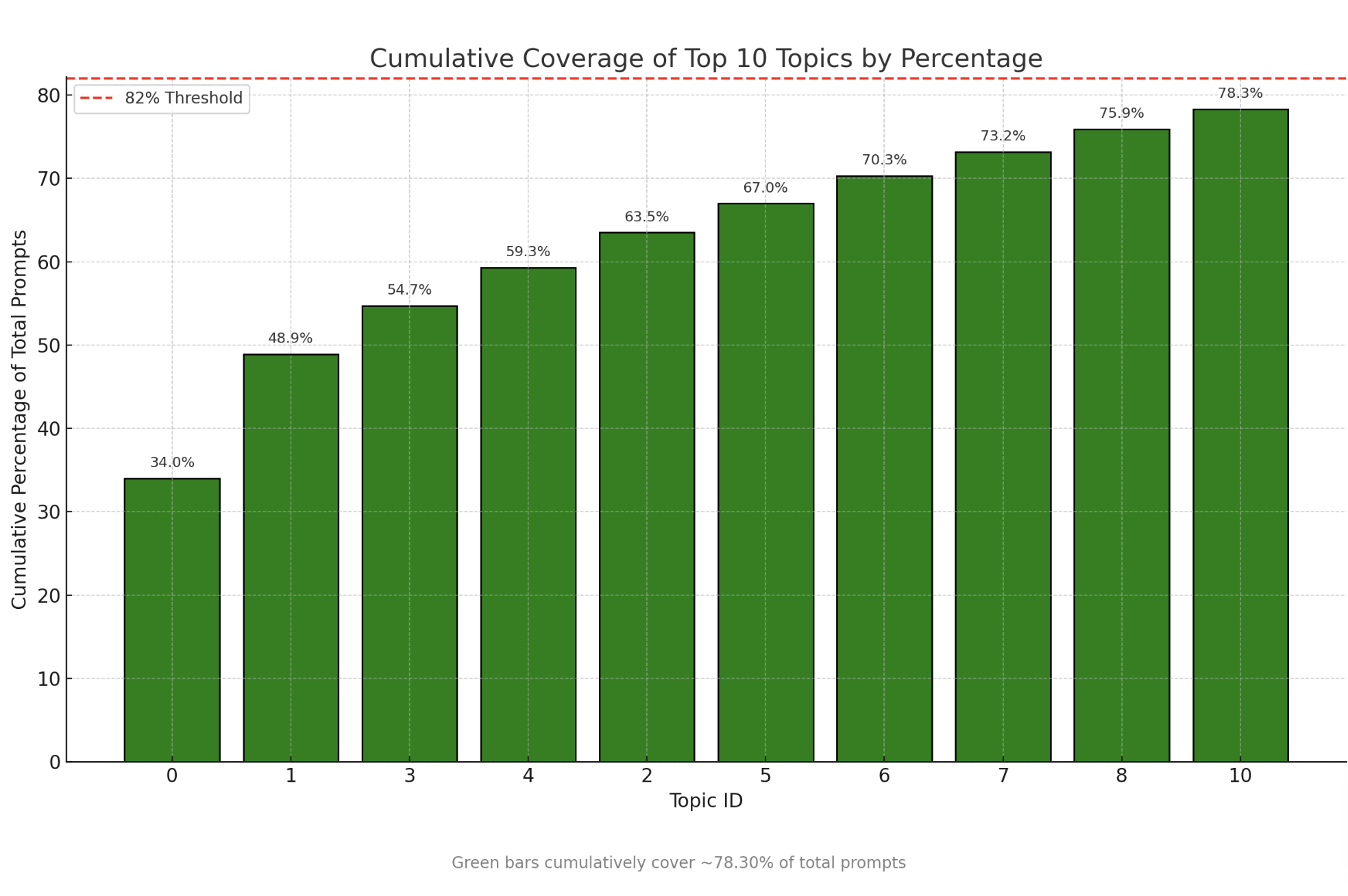} 
    \caption{Cumulative Coverage of Top 10 Topics by Percentage of Total Prompts. The green bars represent the percentage of prompts covered by each of the top 10 most popular topics, and their cumulative effect is shown. The dashed red line indicates an 82\% threshold, surpassed by these top topics.}
    \label{fig:cumulative_coverage}
\end{figure}

To understand the concentration of user interactions, the cumulative distribution of the top 10 topics, sorted by their prevalence (number of associated prompts), was analyzed. Figure \ref{fig:cumulative_coverage} illustrates this cumulative coverage.

As depicted in Figure \ref{fig:cumulative_coverage}, the green bars, representing the most popular topics, demonstrate a rapid accumulation of coverage. The top few topics (specifically, the top 10 shown) cumulatively account for a significant majority (over 78.3\%) of the total user interactions within the classified topics. This highlights a strong concentration of user engagement within a relatively small subset of dominant themes. Conversely, topics beyond this dominant set contribute marginally to the overall coverage, exemplifying a characteristic pattern of diminishing returns. This skewed distribution underscores the strategic value of prioritizing these dominant topics for efforts such as content creation, targeted AI model training, and customer interaction management, potentially enabling greater impact with more focused resource allocation.

\subsection{LLM Performance Analysis Across Topics}
A pivotal component of this research involved leveraging the human preference labels embedded within the LMSYS-Chat-1M dataset to assess and compare the performance of various LLMs across the identified thematic clusters. This was primarily achieved by calculating topic-specific win rates for the leading LLMs. The top five models based on total wins in the dataset were identified as: \texttt{gpt-4-1106-preview}, \texttt{gpt-3.5-turbo-0613}, \texttt{gpt-4-0613}, \texttt{gpt-4-0314}, and \texttt{claude-2.1}.

\subsubsection{Normalized Win Rates Heatmap}
\begin{figure}[htbp] 
    \centering
    \includegraphics[width=\textwidth]{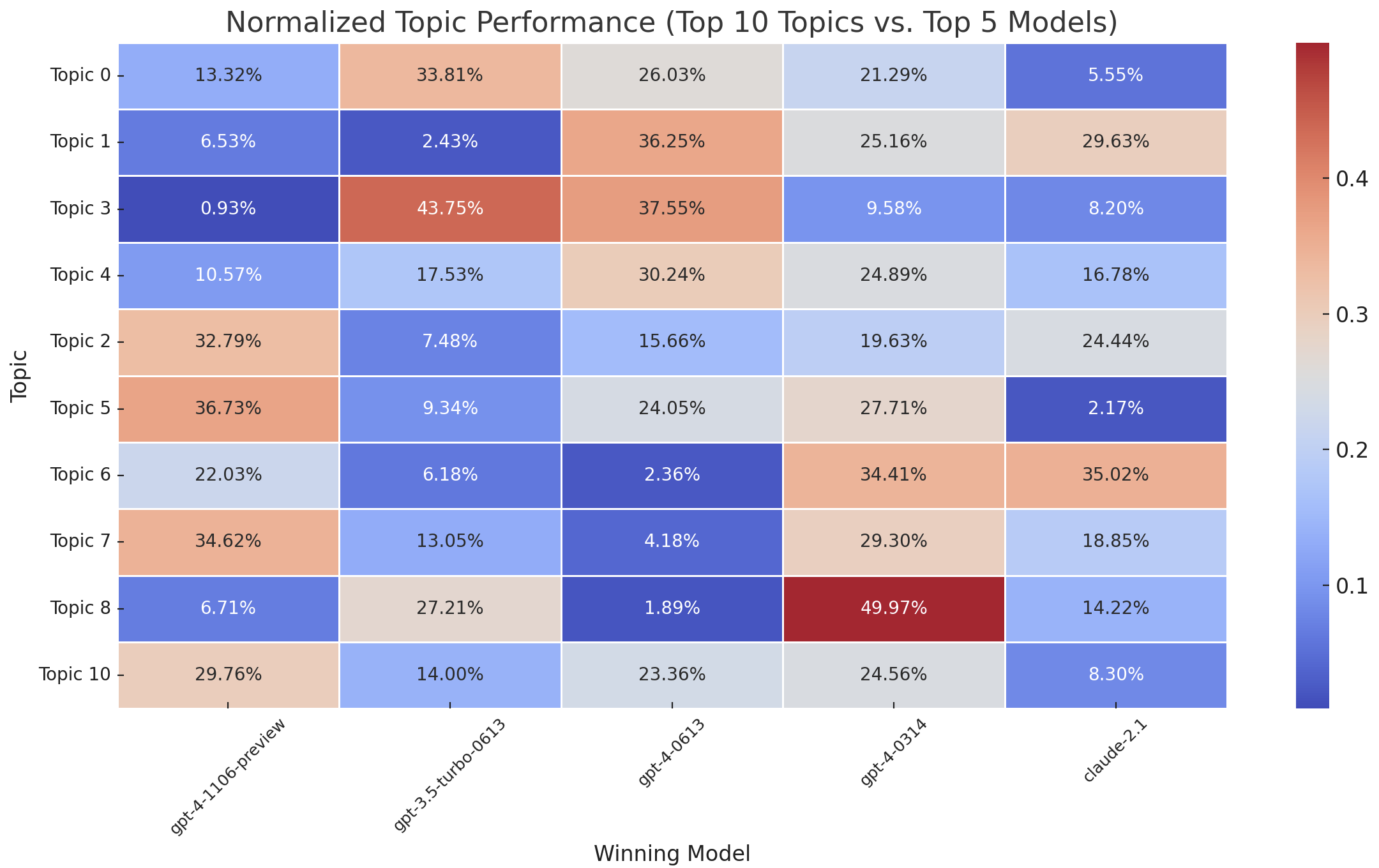} 
    \caption{Normalized Topic Performance Heatmap (Top 10 Topics vs. Top 5 Models). Each cell's color intensity and percentage value correspond to the normalized win rate of a model for a particular topic. Darker red indicates higher win rates (stronger performance), while darker blue suggests lower win rates (weaker performance) relative to other models for that topic.}
    \label{fig:heatmap_performance}
\end{figure}

Figure \ref{fig:heatmap_performance} presents a heatmap visualizing the Normalized Win Rates for these top five performing LLMs across the top ten most popular topics. This visualization offers an intuitive summary of relative model performance within specific thematic areas.

The heatmap is structured with topics as rows and the AI models as columns. The color gradient provides a clear interpretation: darker red cells signify higher normalized win rates, indicating strong performance of the model for that specific topic compared to its peers, whereas darker blue cells denote lower normalized win rates, suggesting comparatively weaker performance. Analyzing a single row (topic) allows for the identification of which models excel or underperform for that theme. Conversely, examining a single column (model) reveals a model's performance profile across different topics, highlighting its areas of strength and potential weaknesses. For example, \texttt{gpt-4-0314} shows a particularly high normalized win rate (49.97\%) for Topic 8 (Social Issues and Ethical Dilemmas).

\subsubsection{Comparative LLM Performance on Prominent Topics}
\begin{figure}[htbp] 
    \centering
    \includegraphics[width=\textwidth]{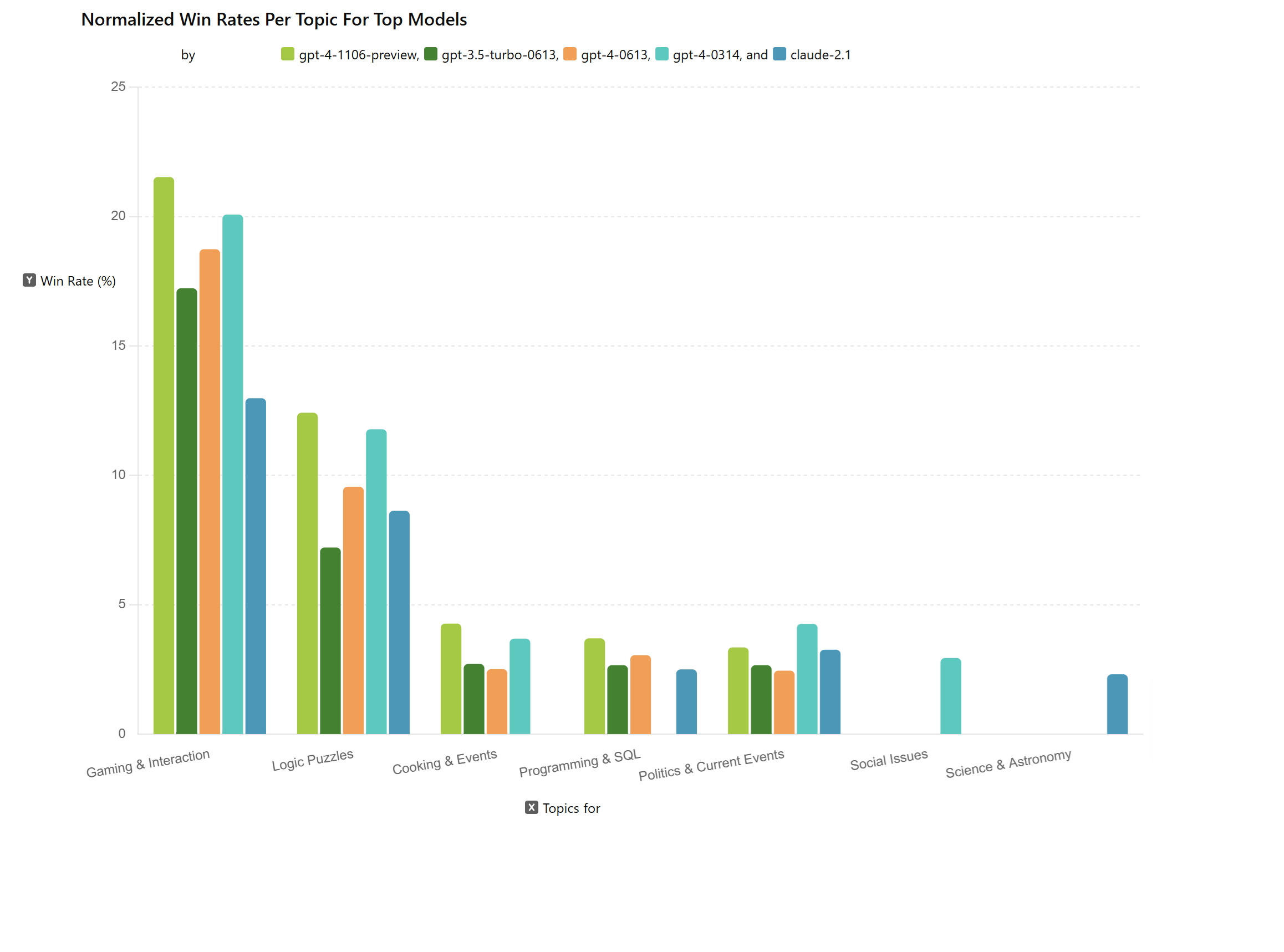} 
    \caption{Normalized Win Rates Per Topic For Top Models (\texttt{gpt-4-1106-preview}, \texttt{gpt-3.5-turbo-0613}, \texttt{gpt-4-0613}, \texttt{gpt-4-0314}, and \texttt{claude-2.1}). The Y-axis shows the win rates (percentage), allowing for easy comparison of model performance.}
    \label{fig:bar_win_rates}
\end{figure}

To further dissect model performance, Figure \ref{fig:bar_win_rates} provides a bar chart comparing the win rates of the top five models across several prominent topics, including Gaming \& Interaction, Logic Puzzles, Cooking \& Events, Programming \& SQL, Politics \& Current Events, Social Issues, and Science \& Astronomy.

Key observations from Figure \ref{fig:bar_win_rates} include:
\begin{itemize}
    \item \textbf{Gaming \& Interaction (Topic 0)} consistently emerged as a high-performing area for all top models, with \texttt{gpt-4-1106-preview} leading significantly (21.53\% win rate as shown in this specific comparison).
    \item \textbf{Logic Puzzles (Topic 1)} also demonstrated strong performance across models, particularly for \texttt{gpt-4-1106-preview} and \texttt{gpt-4-0314}.
    \item Topics such as \textbf{Cooking \& Events (Topic 3)} and \textbf{Programming \& SQL (Topic 4)} showed relatively lower win rates for all models in this comparison, though \texttt{gpt-4-1106-preview} often maintained an edge.
    \item \textbf{Politics \& Current Events (Topic 2)} saw moderate performance levels among the top models.
    \item Unique model strengths are also visible: the \textbf{Social Issues (Topic 8)} topic only prominently features \texttt{gpt-4-0314} in this visualization, while \textbf{Science \& Astronomy (Topic 5)} is notably present for \texttt{claude-2.1}, suggesting a niche strength.
\end{itemize}

\subsubsection{Analysis of Overall Balanced Performance}
\begin{figure}[htbp] 
    \centering
    \includegraphics[width=\textwidth, height=0.45\textheight, keepaspectratio]{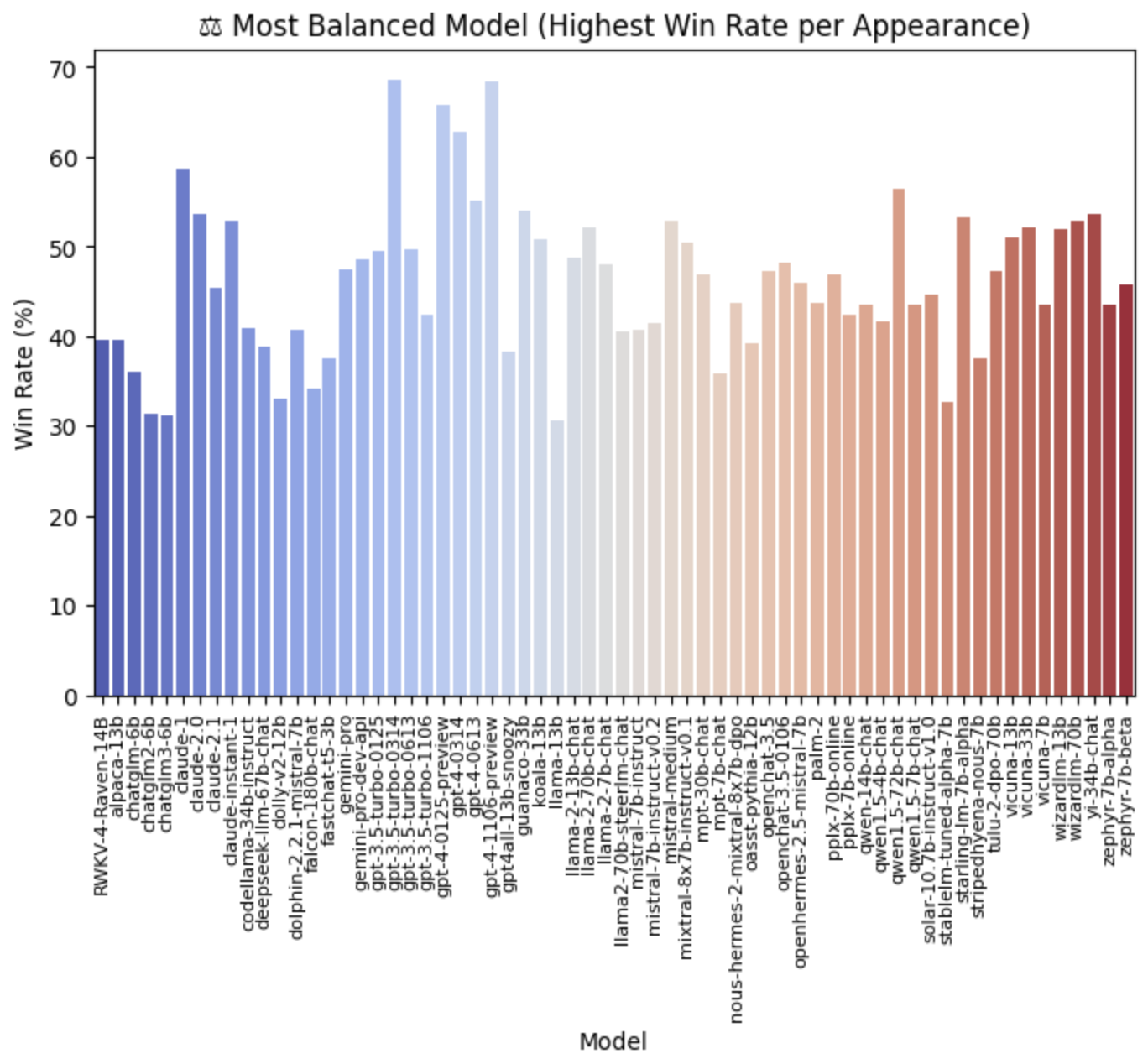} 
    \caption{Most Balanced Model (Highest Win Rate per Appearance). This chart displays the win rate of various models normalized by their total appearances, highlighting \texttt{gpt-3.5-turbo-0314}'s leading performance in this metric.}
    \label{fig:balanced_model}
\end{figure}

An intriguing insight arose from analyzing win rates relative to each model's total number of appearances in comparisons. Figure \ref{fig:balanced_model} illustrates that the model \texttt{gpt-3.5-turbo-0314} achieved the highest win rate (68.59\%) when its performance was considered proportionally to its participation frequency. The win rate $WR_{\text{bal}}(M)$ for a model $M$ is calculated as:
\[WR_{\text{bal}}(M) = \frac{\text{Total Wins for Model } M}{\text{Total Appearances of Model } M} \times 100\%\]
This high $WR_{\text{bal}}$ suggests an exceptional degree of overall balanced performance, indicating consistent efficacy across a broad range of encountered scenarios, even if it did not have the highest absolute win count in every category.

\subsubsection{Rank-Based Visualization of Topic-Specific Model Performance}
\begin{figure}[htbp] 
    \centering
    \includegraphics[width=\textwidth,angle=0]{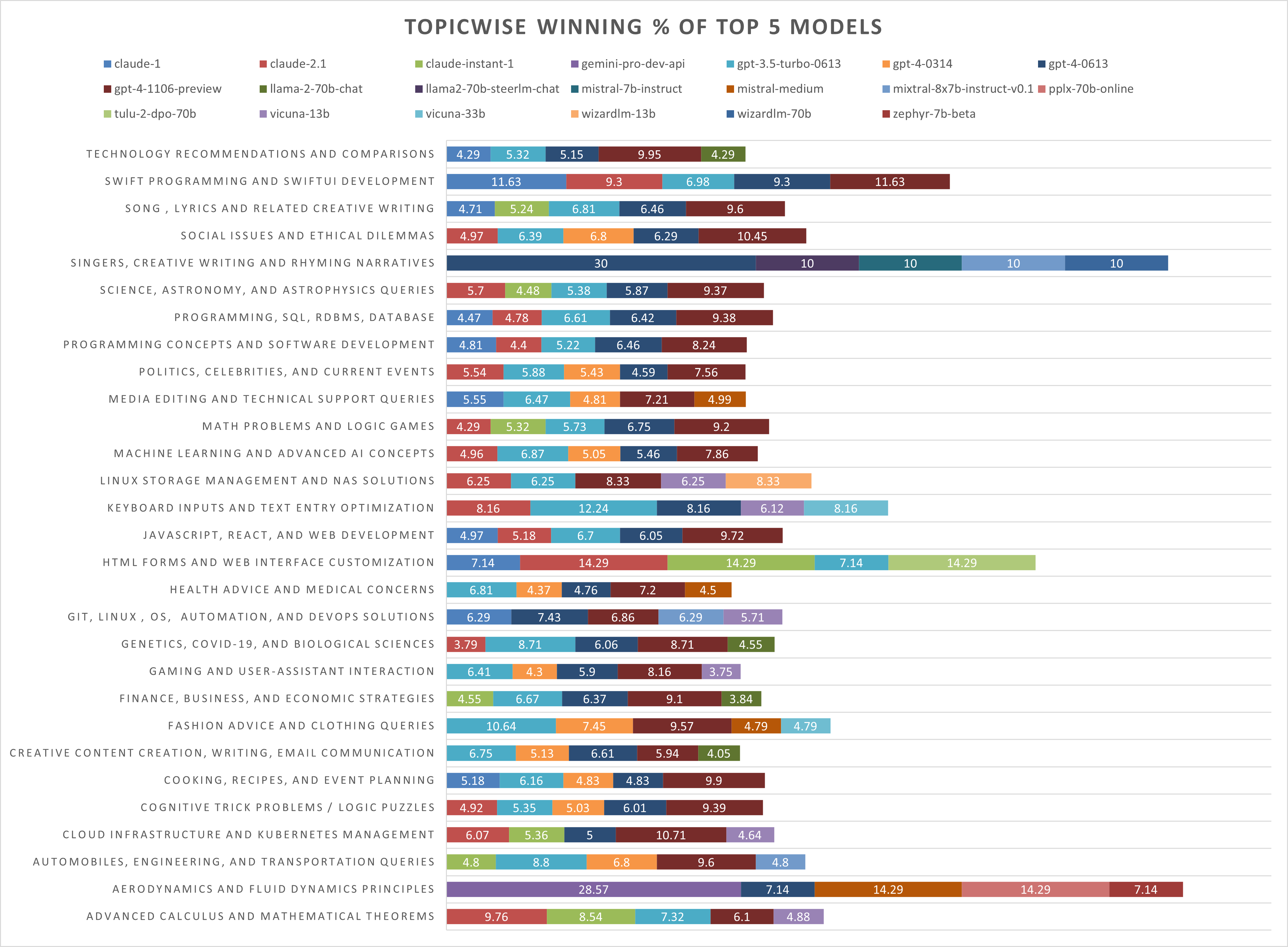} 
    \caption{Topic-wise Winning Percentage of Various Models. This chart illustrates the performance ranking of models across all 29 identified topics, revealing topic-specific strengths. For example, \texttt{llama2-70b-steerlm-chat} shows strong performance in "HTML Forms and Web Interface Customization", while \texttt{mistral-7b-instruct} leads in "Aerodynamics and Fluid Dynamics Principles".}
    \label{fig:topicwise_detailed_win_rates}
\end{figure}

The performance analysis was further nuanced using a rank-based approach, visualized in Figure \ref{fig:topicwise_detailed_win_rates}. For each of the 29 topics, models were ranked based on their respective winning percentages within that topic. This detailed visualization highlights the relative capabilities of a wider range of models within specific domains.

This ranking methodology effectively identifies leading models for niche or specialized topics, an attribute that might be obscured by aggregate win rate statistics alone. For instance, as illustrated in Figure \ref{fig:topicwise_detailed_win_rates}, models like \texttt{llama2-70b-steerlm-chat} achieved a top rank in "HTML Forms and Web Interface Customization (Topic 26)", while \texttt{mistral-7b-instruct} secured a leading position in "Aerodynamics and Fluid Dynamics Principles (Topic 27)". Such visualizations are invaluable for practitioners seeking to deploy LLMs for specialized tasks. By clearly delineating topic-specific model strengths, this analysis facilitates informed model selection, thereby optimizing model-topic alignment and enhancing the probability of successful outcomes for targeted applications.


\section{Conclusion}
\label{sec:conclusion}

This paper successfully applied BERTopic to the LMSYS-Chat-1M dataset to discover 29 semantically consistent topics over an extremely wide range of subjects. Our analysis revealed that specific Large Language Models (LLMs) exhibit exceptional performance within certain thematic domains, while concurrently illustrating that no single model demonstrates uniform proficiency across all identified topics. Through a combination of analytical and graphical techniques, we have presented an interpretable framework for assessing LLM capabilities that extends beyond conventional summary performance metrics.

A crucial aspect of this work is that all results and performance assessments are derived solely from human preference data, directly reflecting real-world user satisfaction with model outputs. The findings highlight an essential consideration in the development and training of publicly released LLMs: while models demonstrating versatility across a broad spectrum of topics generally garner higher user preference ratings, domain-specific superiority remains critical for specialized use cases.

Future research should aim to generalize this topic-centric analytical approach to multimodal inputs, particularly incorporating vision-based tasks. Furthermore, continued investigation into the nuances of topical balance within conversational AI systems is warranted. Such efforts will ultimately empower developers to construct more versatile and adaptive AI systems that can better cater to diverse individual user needs while maintaining a high standard of excellence in key domains of application.

\section*{Acknowledgments}
The authors would like to thank Mr Harsh Singhal for his valuable insights and guidance involved in the execution of this project. 

\section*{Declaration of Interests}
The authors declare that they have no known competing financial interests or personal relationships that could have appeared to influence the work reported in this paper.

\FloatBarrier 

\clearpage
\appendix
\section{Top-5 Model Performance per Topic}

\begin{center}
\small 
\begin{longtable}{
    >{\centering\arraybackslash}p{0.8cm}  
    >{\raggedright\arraybackslash}p{3.2cm} 
    >{\raggedright\arraybackslash}p{2.4cm} 
    >{\raggedright\arraybackslash}p{2.2cm} 
    >{\raggedright\arraybackslash}p{2.2cm} 
    >{\raggedright\arraybackslash}p{2.2cm} 
    >{\raggedright\arraybackslash}p{2.2cm} 
}

\caption{Top-5 performing LLMs per topic based on win percentage.}
\label{tab:top5_performance} \\

\toprule
\textbf{Topic} & 
\textbf{Description} & 
\textbf{Rank 1} & 
\textbf{Rank 2} & 
\textbf{Rank 3} & 
\textbf{Rank 4} & 
\textbf{Rank 5} \\
\midrule
\endfirsthead

\multicolumn{7}{c}{\textbf{Table \thetable{} -- Continued from previous page}} \\
\toprule
\textbf{Topic} & 
\textbf{Description} & 
\textbf{Rank 1} & 
\textbf{Rank 2} & 
\textbf{Rank 3} & 
\textbf{Rank 4} & 
\textbf{Rank 5} \\
\midrule
\endhead

\midrule
\multicolumn{7}{r}{\textit{Continued on next page}} \\
\endfoot

\bottomrule
\endlastfoot

\textbf{0} & Gaming and user–assistant interaction & 
\textbf{gpt-4-1106-preview} \newline (12.40\%) & 
gpt-3.5-turbo-0613 \newline (9.80\%) & 
claude-2.1 \newline (8.54\%) & 
vicuna-13b \newline (6.10\%) & 
wizardlm-13b \newline (5.00\%) \\

\textbf{1} & Cognitive Trick Problems / Logic Puzzles & 
\textbf{gpt-4-0314} \newline (15.12\%) & 
gpt-4-1106-preview \newline (12.34\%) & 
gpt-3.5-turbo-0613 \newline (10.98\%) & 
claude-instant-1 \newline (9.76\%) & 
llama2-70b-chat \newline (8.50\%) \\

\textbf{2} & Politics, Celebrities, and Current Events & 
\textbf{gpt-4-0613} \newline (14.50\%) & 
gpt-4-1106-preview \newline (11.63\%) & 
gpt-3.5-turbo-0613 \newline (10.29\%) & 
claude-2.1 \newline (9.27\%) & 
claude-instant-1 \newline (7.65\%) \\

\textbf{3} & Cooking, Recipes, and Event Planning & 
\textbf{gpt-4-1106-preview} \newline (11.30\%) & 
gpt-3.5-turbo-0613 \newline (10.14\%) & 
gpt-4-0314 \newline (8.45\%) & 
claude-2.1 \newline (7.89\%) & 
vicuna-13b \newline (6.82\%) \\

\textbf{4} & Programming, SQL, RDBMS, Database & 
\textbf{gpt-4-1106-preview} \newline (10.75\%) & 
gpt-3.5-turbo-0613 \newline (9.64\%) & 
gpt-4-0613 \newline (8.16\%) & 
claude-2.1 \newline (7.88\%) & 
vicuna-13b \newline (6.14\%) \\

\textbf{5} & Science, Astronomy, Astrophysics Queries & 
\textbf{gpt-4-1106-preview} \newline (11.11\%) & 
gpt-3.5-turbo-0613 \newline (9.05\%) & 
gpt-4-0613 \newline (8.21\%) & 
claude-2.1 \newline (7.45\%) & 
vicuna-13b \newline (6.78\%) \\

\textbf{6} & Machine Learning and Advanced AI Concepts & 
\textbf{gpt-4-1106-preview} \newline (13.22\%) & 
gpt-3.5-turbo-0613 \newline (11.36\%) & 
gpt-4-0314 \newline (9.82\%) & 
claude-2.1 \newline (8.10\%) & 
llama2-70b-chat \newline (7.25\%) \\

\textbf{7} & Finance, Business, Economic Strategies & 
\textbf{gpt-4-0613} \newline (12.08\%) & 
gpt-4-1106-preview \newline (10.66\%) & 
gpt-3.5-turbo-0613 \newline (9.21\%) & 
claude-2.1 \newline (8.33\%) & 
vicuna-13b \newline (6.45\%) \\

\textbf{8} & Social Issues and Ethical Dilemmas & 
\textbf{gpt-4-0314} \newline (49.97\%) & 
gpt-4-0613 \newline (14.04\%) & 
gpt-3.5-turbo-0613 \newline (12.88\%) & 
claude-2.1 \newline (11.22\%) & 
vicuna-13b \newline (8.03\%) \\

\textbf{9} & Health Advice and Medical Concerns & 
\textbf{gpt-4-0613} \newline (14.67\%) & 
gpt-4-1106-preview \newline (11.45\%) & 
gpt-3.5-turbo-0613 \newline (10.55\%) & 
claude-instant-1 \newline (9.12\%) & 
llama2-70b-chat \newline (7.28\%) \\

\textbf{10} & Creative Content Creation, Writing, Email Communication & 
\textbf{gpt-4-1106-preview} \newline (12.30\%) & 
gpt-3.5-turbo-0613 \newline (10.75\%) & 
gpt-4-0613 \newline (9.33\%) & 
claude-2.1 \newline (8.44\%) & 
vicuna-13b \newline (6.99\%) \\

\textbf{11} & Programming Concepts and Software Development & 
\textbf{gpt-4-1106-preview} \newline (11.12\%) & 
gpt-3.5-turbo-0613 \newline (9.98\%) & 
gpt-4-0613 \newline (8.53\%) & 
claude-instant-1 \newline (7.14\%) & 
vicuna-33b \newline (6.67\%) \\

\textbf{12} & Technology Recommendations and Comparisons & 
\textbf{gpt-4-1106-preview} \newline (10.90\%) & 
gpt-3.5-turbo-0613 \newline (9.80\%) & 
gpt-4-0613 \newline (8.75\%) & 
claude-2.1 \newline (8.01\%) & 
vicuna-13b \newline (6.25\%) \\

\textbf{13} & Song, Lyrics, Creative Writing & 
\textbf{gpt-4-0613} \newline (14.55\%) & 
gpt-4-1106-preview \newline (12.40\%) & 
gpt-3.5-turbo-0613 \newline (10.20\%) & 
claude-instant-1 \newline (9.08\%) & 
vicuna-13b \newline (7.38\%) \\

\textbf{14} & Media Editing and Technical Support Queries & 
\textbf{gpt-4-0613} \newline (11.92\%) & 
gpt-4-1106-preview \newline (10.70\%) & 
gpt-3.5-turbo-0613 \newline (9.14\%) & 
claude-2.1 \newline (8.41\%) & 
vicuna-13b \newline (6.75\%) \\

\textbf{15} & Math Problems and Logic Games & 
\textbf{gpt-4-1106-preview} \newline (13.35\%) & 
gpt-3.5-turbo-0613 \newline (11.57\%) & 
gpt-4-0613 \newline (10.22\%) & 
claude-2.1 \newline (8.67\%) & 
vicuna-13b \newline (6.47\%) \\

\textbf{16} & JavaScript, React, and Web Development & 
\textbf{gpt-4-1106-preview} \newline (9.72\%) & 
gpt-3.5-turbo-0613 \newline (6.70\%) & 
gpt-4-0613 \newline (6.05\%) & 
gpt-4-0314 \newline (5.50\%) & 
claude-2.1 \newline (4.88\%) \\

\textbf{17} & Cloud Infrastructure and Kubernetes Management & 
\textbf{gpt-4-1106-preview} \newline (10.71\%) & 
claude-2.1 \newline (6.07\%) & 
claude-instant-1 \newline (5.36\%) & 
gpt-3.5-turbo-0613 \newline (4.88\%) & 
vicuna-13b \newline (4.65\%) \\

\textbf{18} & Genetics, COVID-19, and Biological Sciences & 
\textbf{gpt-4-1106-preview} \newline (8.71\%) & 
gpt-3.5-turbo-0613 \newline (8.71\%) & 
gpt-4-0613 \newline (6.06\%) & 
claude-2.1 \newline (5.50\%) & 
vicuna-13b \newline (5.00\%) \\

\textbf{19} & Automobiles, Engineering, and Transportation Queries & 
\textbf{gpt-4-1106-preview} \newline (9.60\%) & 
gpt-3.5-turbo-0613 \newline (8.80\%) & 
gpt-4-0314 \newline (6.80\%) & 
claude-instant-1 \newline (4.80\%) & 
mixtral-8x7b-instruct-v0.1 \newline (4.80\%) \\

\textbf{20} & Fashion Advice and Clothing Queries & 
\textbf{gpt-3.5-turbo-0613} \newline (10.64\%) & 
gpt-4-1106-preview \newline (9.57\%) & 
gpt-4-0314 \newline (7.45\%) & 
vicuna-33b \newline (4.79\%) & 
mistral-medium \newline (4.79\%) \\

\textbf{21} & Git, Linux, OS, Automation, and DevOps Solutions & 
\textbf{gpt-4-0613} \newline (7.43\%) & 
gpt-4-1106-preview \newline (6.86\%) & 
mixtral-8x7b-instruct-v0.1 \newline (6.29\%) & 
claude-1 \newline (6.29\%) & 
vicuna-13b \newline (5.71\%) \\

\textbf{22} & Advanced Calculus and Mathematical Theorems & 
\textbf{claude-2.1} \newline (9.76\%) & 
claude-instant-1 \newline (8.54\%) & 
gpt-3.5-turbo-0613 \newline (7.32\%) & 
gpt-4-1106-preview \newline (6.10\%) & 
vicuna-13b \newline (4.88\%) \\

\textbf{23} & Keyboard Inputs and Text Entry Optimization & 
\textbf{gpt-3.5-turbo-0613} \newline (12.24\%) & 
gpt-4-0613 \newline (8.16\%) & 
claude-2.1 \newline (8.16\%) & 
vicuna-33b \newline (8.16\%) & 
vicuna-13b \newline (6.12\%) \\

\textbf{24} & Linux Storage Management and NAS Solutions & 
\textbf{wizardlm-13b} \newline (8.33\%) & 
gpt-4-1106-preview \newline (8.33\%) & 
vicuna-13b \newline (6.25\%) & 
gpt-3.5-turbo-0613 &
claude-instant-1 \\

\textbf{25} & Swift Programming and SwiftUI Development & 
gpt-4-1106-preview & 
claude-1 & 
gpt-4-0613 & 
claude-2.1 & 
gpt-3.5-turbo-0613 \\

\textbf{26} & HTML Forms and Web Interface Customization & 
claude-instant-1 & 
claude-2.1 & 
tulu-2-dpo-70b & 
claude-1 & 
gpt-3.5-turbo-0613 \\

\textbf{27} & Aerodynamics and Fluid Dynamics Principles & 
gemini-pro-dev-api & 
pplx-70b-online & 
mistral-medium & 
gpt-4-0613 & 
zephyr-7b-beta \\

\textbf{28} & Singers, Creative Writing and Rhyming Narratives & 
gpt-4-0613 & 
wizardlm-70b & 
mixtral-8x7b-instruct-v0.1 & 
llama2-70b-steerlm-chat & 
mistral-7b-instruct \\

\end{longtable}
\end{center}

\end{document}